\documentclass{ecai} 

\usepackage{latexsym}
\usepackage{amssymb}
\usepackage{amsmath}
\usepackage{amsthm}
\usepackage{booktabs}
\usepackage{enumitem}
\usepackage{graphicx}
\usepackage{color}

\usepackage{hyperref}

\usepackage{tabularx}

\newcommand{\BibTeX}{B\kern-.05em{\sc i\kern-.025em b}\kern-.08em\TeX}
\newcommand{\imgcountpre}{483,840}
\newcommand{\imgcountfinal}{246,285}
\newcommand{\promptcount}{96,768}

\begin{document}


\begin{frontmatter}


\paperid{6722} 


\title{On Synthetic Texture Datasets: Challenges, Creation, and Curation}


\author[A]{\fnms{Blaine}~\snm{Hoak}\orcid{0000-0003-2960-0686}\thanks{Corresponding Author. Email: bhoak@wisc.edu}}
\author[A]{\fnms{Patrick}~\snm{McDaniel}\orcid{0000-0003-2091-7484}} 

\address[A]{University of Wisconsin-Madison}


\begin{abstract}
Texture data serves as a valuable tool for interpreting the high-level features models learn, uncovering biases, and identifying security vulnerabilities. However, works in this space have been limited by small texture datasets and synthesis methods that struggle to scale in the diversity and specificity required for these tasks. In this work, we introduce an extensible methodology for generating high-quality, diverse texture images, which we use to create the Prompted Textures Dataset (PTD), a new texture dataset spanning \imgcountfinal{} images across 56 texture classes. Our comparison against real texture data demonstrates that PTD is more diverse while maintaining quality. Additionally, human evaluations confirm that every stage in our methodology enhances texture quality, yielding a 3.4\% increase in quality and a 4.5\% increase in representativeness overall. Our dataset is available for download at \url{https://zenodo.org/records/15359142}.
\end{abstract}

\end{frontmatter}


\section{Introduction}

Large, high-quality datasets have driven advancements across diverse AI fields, including object classification, visual emotion recognition, medical image interpretation, scene recognition, and more~\cite{russakovsky_imagenet_2015, you_building_2016, irvin_chexpert_2019, zhou_places_2018}. Texture data, in particular, is essential for understanding the high-level features models learn and their implications. Studies have shown that models exhibit texture bias~\cite{geirhos_imagenet-trained_2019}, that texture data can aid in constructing texture-object associations~\cite{hoak_explorations_2024}, and that textures can even be exploited to create adversarial examples~\cite{zhang_practical_2022}.

However, texture-based research has been constrained by the limited availability of diverse and scalable texture datasets. Existing datasets, often rely on manual image collection, typically from public sources like Flickr~\cite{fearon_263365_2014}, resulting in small, specialized sets of images that limit the scope of broader texture analysis. Consequently, most studies rely on datasets with \textit{fewer than 100 texture images}, making it challenging to conduct large-scale analyses or evaluate the generalization capabilities of models on texture data and necessitating methods for synthesizing new textures.

Furthermore, traditional texture synthesis methods are often example based, meaning they rely on pre-existing texture examples as seed images, which still does not remove the burden of manual image collection, and restricts the diversity and scalability of the resulting dataset~\cite{gatys_texture_2015, gatys_neural_2015, portilla_parametric_2000, zhou_non-stationary_2018}. Additionally, since these methods operate on an image-to-image basis, the resulting images lack textual descriptions, making it challenging to control for specific texture characteristics or align textures with semantic labels. This limitation becomes particularly restrictive when textures must be generated to match specific classes or descriptors~\cite{geirhos_imagenet-trained_2019, hoak_explorations_2024}. 

\begin{figure}[t]
  \includegraphics[width=\linewidth]{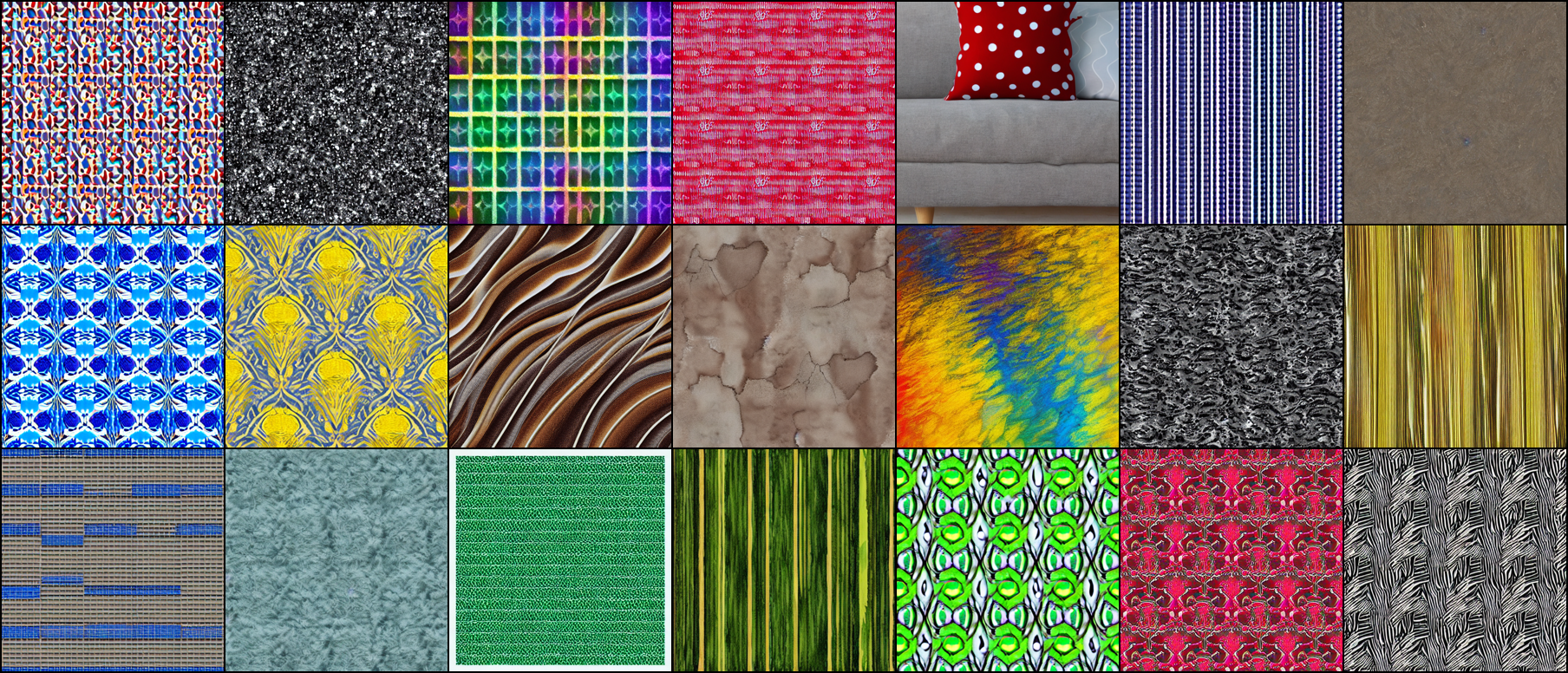}
  \caption{Prompted Textures Dataset (PTD) (our work).}
  \label{fig:ptd_images}
  \vspace{0.3cm}
\end{figure}

\begin{figure}[t]
  \includegraphics[width=\linewidth]{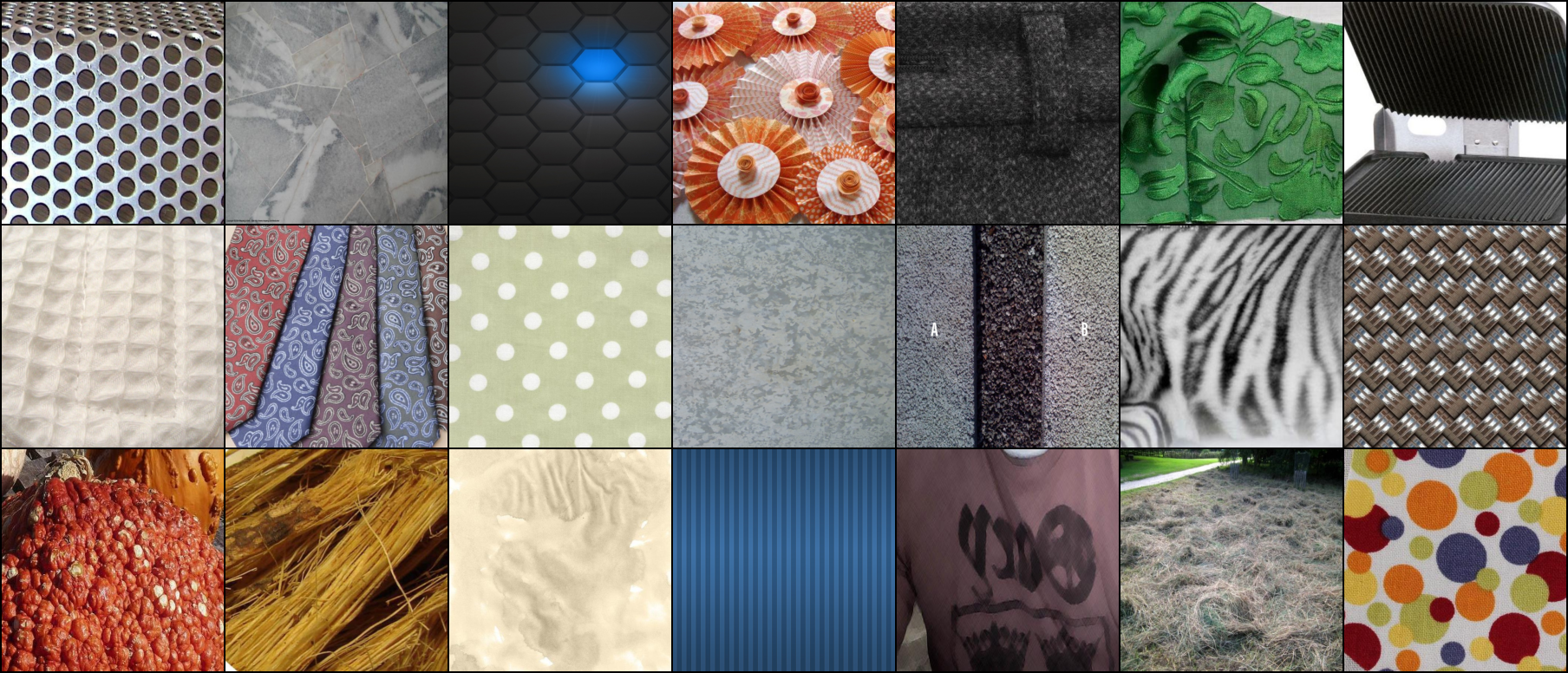}
  \caption{Describable Textures Dataset (DTD)\protect\cite{cimpoi_describing_2014}.}
  \label{fig:dtd_images}
  \vspace{0.3cm}
\end{figure}

In this work, we leverage text-to-image models to introduce a new, extensible
methodology for generating high-quality, diverse, and specific texture images
capable of supporting a broad range of texture-based tasks. Here, we translate
descriptive prompts into visually representative texture images by adapting
traditional text-to-image pipelines with texture-specific considerations. These
adaptations allow us to create the Prompted Textures Dataset (PTD), a dataset of
\imgcountfinal{} texture images across 56 texture classes. Examples of the PTD
are shown in \autoref{fig:ptd_images}, alongside real texture data from the Describable Textures Dataset
(DTD)~\cite{cimpoi_describing_2014}, in \autoref{fig:dtd_images}.

Our approach takes place in three steps. First, we produce texture-specific prompts to serve as the basis for our texture generation, where we apply combinations of artistically-informed descriptors across a wide range of texture classes to enable controlled diversity of textures. Second, we use our constructed prompts as inputs to Stable Diffusion model pipelines, which we adapt based on unique challenges we uncover that arise from safety filters sensitivity to texture data, to produce the corresponding texture images. Finally, to yield higher-quality and more texture-like images, we perform a three stage refinement process consisting of frequency analysis, patch variance filtering, and CLIP scoring to filter down the images.

To validate the PTD, we conduct a comparison against real texture data and find that our dataset is high-quality, diverse, and representative of real textures. We additionally conduct a human evaluation on our data at each stage of our refinement process resulting in a 3.4\% and 4.5\% increase in quality and representativeness, respectfully, with each stage of our refinement process positively contributing to the overall success. Finally, we analyze trends in prompts that yield the best texture images.

As an additional validation, we analyze the \textit{Texture Object Association Values (TAV)}~\cite{hoak_err_2024} formed by using the Prompted Textures Dataset. TAV is a new metric that has been introduced since the public release of our dataset that leverages the Prompted Textures Dataset (introduced in this work) to uncover the associations between learned textures and objcts in object classification models for the purposes of measuring texture bias.

\section{Background}\label{sec:background}

\subsection{Texture datasets}\label{sec:background:datasets}
The Describable Textures Dataset (DTD) \cite{cimpoi_describing_2014} is perhaps the
most popular texture dataset to date. It contains 5640 images sourced from
Flickr in 47 texture categories such as polka-dotted, scaly, and striped. This
texture dataset has had a variety of uses in computer vision and machine
learning. 
Aside from the DTD, other works operating on texture datasets have created their
own sets of textures to suit their specific use cases.
In \cite{geirhos_imagenet-trained_2019}, the authors construct a shape-cue
conflict dataset, which contains images with the texture of one object and the
shape of another for the purposes of studying if CNNs were more biased towards
texture or shape.  
Finally, in \cite{zhang_practical_2022} patterned images were created and
overlayed onto existing object images; the authors found that this method produced an
effective attack against machine learning models, wherein these patterns caused
the model to misclassify the images and could be constructed even without any
access to the model weights or training data.

\subsection{Texture synthesis}\label{sec:background:synthesis}


\subsubsection{Classical Methods}
Classical texture synthesis methods typically use a sample texture as a reference, generating new patterns by sampling or modeling its visual characteristics. Early non-parametric approaches, such as pixel-based synthesis~\cite{efros_texture_1999}, generate textures by matching local neighborhoods to capture fine-grained details. Patch-based methods like image quilting~\cite{efros_image_2001} improve texture coherence by stitching larger patches, reducing visible seams in simpler textures. Statistical models introduced additional flexibility by matching statistical properties of texture features. One of the earliest examples introduced a parametric model that synthesizes textures by matching wavelet coefficient statistics, making it suitable for stationary textures~\cite{portilla_parametric_2000}. Later advancements in methods like texture stationarization~\cite{Moritz2017Texture} and repeatable pattern extraction~\cite{rodriguez-pardo_automatic_2019} further improved tileability and pattern regularity. However, these methods assume uniformity and struggle with complex or large-scale structures.


\subsubsection{Deep learning based methods}
The advent of deep learning introduced powerful tools for texture synthesis, enabling more flexible and complex generation processes. Convolutional neural networks (CNNs) have been pivotal in this shift, with~\citet{gatys_texture_2015} pioneering the use of CNNs to capture textural information through Gram matrices of feature maps, which laid the foundation for neural style transfer~\cite{gatys_image_2016}. Due to its computationally expensive optimization, \citet{ustyuzhaninov_texture_2016} showed that even shallow networks with random filters could capture essential texture patterns, broadening CNN applications in texture synthesis. Later works introduced faster alternatives, training feed-forward networks to approximate this process in a single pass~\cite{ulyanov_texture_2016, johnson_perceptual_2016}.

Generative Adversarial Networks (GANs) further advanced texture synthesis by introducing adversarial training, where a discriminator evaluates texture realism. For instance, Markovian GANs (MGANs)~\cite{li_precomputed_2016} use patch-based discriminators to ensure local coherence, while PSGANs~\cite{bergmann_learning_2017} employ periodic functions to synthesize high-quality periodic textures. GANs generally struggle with non-stationary textures, but approaches like non-stationary texture synthesis~\cite{zhou_non-stationary_2018} incorporate spatially adaptive normalization, allowing for complex, large-scale structures. Additionally, SeamlessGAN~\cite{rodriguez-pardo_seamlessgan_2022} introduced a self-supervised approach to generate tileable texture maps from a single exemplar, enabling textures with seamless continuity.

Despite these advancements, these methods remain constrained by the need for starting examples of textures. Towards methods to alleviate this burden, newer generative models trained on single images, such as SinGAN~\cite{shaham_singan_2019}, demonstrated that textures can be synthesized from a single exemplar, producing outputs with similar texture characteristics to the input, similar in spirit to style transfer methods. Recent advances in text-driven synthesis, such as Text2Tex~\cite{chen_text2tex_2023}, use diffusion models to generate textures directly from descriptive text prompts, marking a shift from example-based to prompt-driven generation. However, Text2Tex has been tailored specifically for 3D meshes. To the best of our knowledge, there has not yet been any work done on creating a 2D texture dataset from prompts alone.

\subsection{Text-to-image models and data metrics.}\label{sec:background:metrics}

Text-to-image models are a class of generative models that transform textual descriptions into representative images. Among these, Stable Diffusion (SD) has become a leading model, capable of generating high-quality images that align closely with input prompts~\cite{rombach_high-resolution_2022}. These models are trained by progressively adding noise to latent representations of images and learning a denoising process to recover them. During inference, SD transforms random noise into coherent images, guided by text descriptions.

The quality, diversity, and representativeness of generated images can be assessed using several key metrics. CLIP scores~\cite{hessel_clipscore_2022} measure representativeness by calculating the cosine similarity between image and text embeddings from a pre-trained CLIP model. Originally developed for captioning quality, CLIP scores are now widely used in text-to-image evaluation, with higher scores indicating stronger alignment between images and their textual prompts.

Inception Scores~\cite{salimans_improved_2016} evaluate image quality and diversity by measuring the KL divergence between conditional and marginal class probabilities using a pre-trained Inception model. High Inception Scores indicate that generated images are both distinct and strongly predicted as belonging to specific classes, with predictions spread across categories for diverse data.

FID scores~\cite{bynagari_gans_2019} assess realism by calculating the Fréchet distance between feature distributions of generated images and a set of real images, typically from a source dataset. Here, we use the Describable Textures Dataset (DTD)~\cite{cimpoi_describing_2014} as the reference. Lower FID scores suggest greater similarity to real images, indicating that generated images exhibit realistic texture and quality.

\section{The Prompted Textures Dataset (PTD)} 

Here, we introduce our methodology for creating high quality and diverse texture data, and how we apply this methodology to create the Prompted Textures Dataset (PTD).


\subsection{Creating Prompts}\label{sec:prompts}

To generate texture data using text-to-image models, we first construct
prompts that describe the textures we aim to create. Later on, these prompts are input to Stable Diffusion \cite{rombach_high-resolution_2022} to produce the corresponding images.

\subsubsection{Selecting Descriptors.}
Our goal is to create diverse, high-quality prompts that yield a wide variety of texture images. To achieve this, we incorporate descriptors that specify not only texture type but also various attributes like color, style, and pattern structure. This approach allows us to go beyond basic descriptions (e.g., "a striped image") and generate varied representations within each texture class, ensuring controlled diversity rather than relying solely on model randomness.

We begin with the 47 texture classes from the Describable Textures
Dataset~\cite{cimpoi_describing_2014} and expand this list by identifying
additional texture candidates by sourcing additional lists of
textures~\cite{barnett_400_2023}, prioritizing those that are meaningfully
different from our starting textures. From this, we add 9 new texture classes,
resulting in a total of 56 texture classes for our prompts. To enrich the
prompts further, we introduce additional descriptive categories inspired by the
7 basic elements of art: line, shape, form, texture, space, color, and
value~\cite{ocvirk2001art}. We prioritize creating categories that enhance
variation in multiple elements without overlapping with the core texture
classes, keeping the prompt space manageable. We select a few distinctive words
within each category to maximize unique texture representations.

\autoref{tab:magic_words} presents the descriptor categories along with the list of words used in each. Prompts are structured by combining one word from each category in standard English adjective order:

\noindent\textit{
\{Artistic\} \{Spatial\} \{Color Enhancer\} \{Color\} \{Texture\}
}

\noindent This enumeration produces prompts such as “photorealistic randomized vivid red polka-dotted texture,” resulting in \promptcount{} unique prompts for generating our texture images.


\begin{table}[t]
\centering
\begin{tabularx}{\columnwidth}{lX}
\toprule
\textbf{Categories} & \textbf{Descriptors} \\
\midrule
\textbf{textures} & banded, blotchy, braided, bubbly, bumpy, checkered,
cobwebbed, cracked, crosshatched, crystalline, dotted, fibrous, flecked,
freckled, frilly, gauzy, grid, grooved, honeycombed, interlaced, knitted,
lacelike, lined, marbled, matted, meshed, paisley, perforated, pitted, pleated,
polka-dotted, porous, potholed, scaly, smeared, spiraled, sprinkled, stained,
stratified, striped, studded, swirly, veined, waffled, woven, wrinkled,
zigzagged, flaky, chapped, hairy, leathery, feathered, spiky, fluffy, ribbed,
wavy \\
\midrule
\textbf{artistic} & $\emptyset$, impressionist, photorealistic, minimal \\
\midrule
\textbf{spatial} & $\emptyset$, randomized, symmetrical \\
\midrule
\textbf{enhancer} & $\emptyset$, gradient, vivid, muted, iridescent, neon,
faded, watercolor, earthy \\
\midrule
\textbf{color} & $\emptyset$, red, green, blue, yellow, black-and-white, pastel,
neutral \\
\bottomrule
\end{tabularx}

\caption{Descriptors for texture prompts. $\emptyset$ indicates an empty string.}
\label{tab:magic_words}
\end{table}

This prompt creation methodology is versatile and can be adapted to other tasks in image generation. For example, replacing texture descriptors with shape descriptors enables prompts like “photorealistic vivid red circle” to produce shape images. Additionally, this approach could be tailored to generate texture phenomena aligned with specific needs, such as introducing “elephant skin” or “wood grain” textures for studies like those in~\cite{geirhos_imagenet-trained_2019}, where textures are chosen based on their likeness to ImageNet~\cite{russakovsky_imagenet_2015} object classes. By building our pipeline with extensibility in mind, we ensure it can support a wide range of texture-based tasks.

\subsection{Generating Images}\label{sec:gen_images}

To generate images for our dataset, we use our texture prompts as input to the Stable Diffusion model~\cite{rombach_high-resolution_2022}. Stable Diffusion, in addition to the main diffusion component, includes a content safety filter that flags images as NSFW (Not Safe For Work) if they exceed a threshold for CLIP scores with secret NSFW content words (though there have been efforts to reverse engineer the words \cite{rando_red-teaming_2022}). This filtering process replaces flagged images with black screens, a safeguard against potentially inappropriate content.

A notable challenge during image generation was achieving images that passed the NSFW filter, despite using benign prompts. To ensure a consistent number of images, we regenerated images flagged by the filter. For analysis, we modified the pipeline to disable the filter temporarily, allowing us to record flagged images while retaining the original, unfiltered content. Although we \textbf{find no images that actually represent explicit content}, for ethical reasons we exclude these flagged images from our final dataset and do not release them publicly. After generating 5 (not flagged) images per prompt, we have a total of \imgcountpre{} images before additional refinement.

\subsubsection{Investigating Safety Filtering.}\label{sec:filtering}

\begin{figure}[t]
  \includegraphics[width=\linewidth]{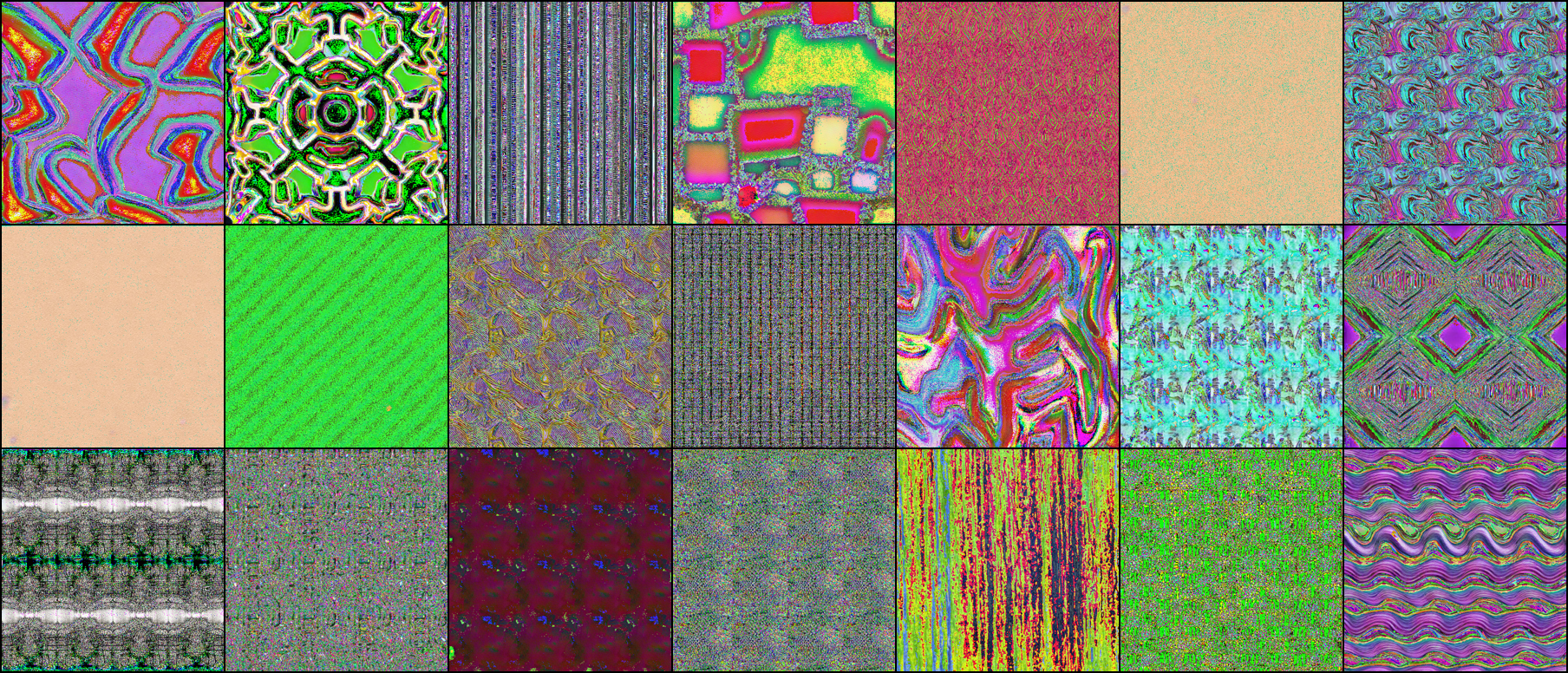}
  \caption{Examples of images flagged as NSFW.}
  \vspace{0.4cm}
  \label{fig:nsfw_images}
\end{figure}

This filtering issue proved substantial, with up to 60\% of our initial generated images being flagged as NSFW. To better understand the filtering process, we examined both flagged and unflagged images. Flagged images typically appeared smoother and more muted in color but did not show clear indicators of explicit content. Examples of flagged images are shown in \autoref{fig:nsfw_images}.

\begin{figure*}[t]
  \includegraphics[width=\textwidth]{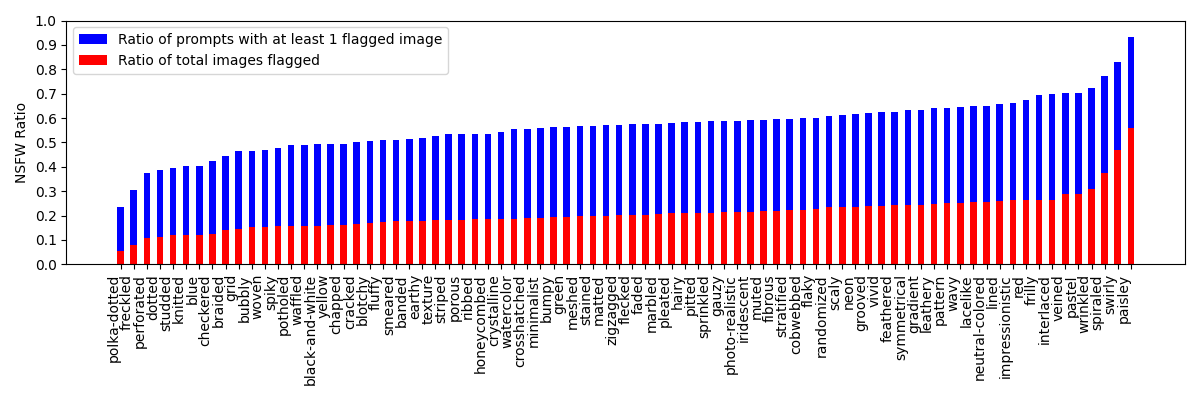}
  \caption{Ratio of total images flagged as NSFW (red) and ratio of prompts with at least one flagged image (blue), organized by word present in the prompt.}
  \label{fig:nsfw_flagging_stacked}
\end{figure*}

We further analyzed NSFW flagging patterns by prompt descriptors. \autoref{fig:nsfw_flagging_stacked} displays the proportion of flagged images and prompts, organized by descriptor. Surprisingly, texture descriptors like “paisley” consistently triggered the filter, with nearly 100\% of prompts containing “paisley” producing at least one flagged image (out of five total generated per prompt). Additionally, many of the top prompt descriptors that led to high flagging rates were \textit{the texture classes themselves}. This indicates that the high NSFW flagging rates are a broader issue with generating texture images, rather than being limited to specific prompts.

While we found no explicit content in flagged images, we excluded these images from further analysis for ethical reasons and did not release them as part of our dataset. This experience highlights the need for refined NSFW filters that can better handle abstract or texture-based content.

\subsection{Refinement Process}\label{sec:refinement}

Since Stable Diffusion models are not explicitly designed to create textures, other kinds of images (e.g., objects) may arise during the generation process. To ensure that our Prompted Textures Dataset (PTD) maintains high quality, diversity, and alignment with prompts, we apply a multi-stage refinement process. This process consists of three filtering steps, each targeting different aspects of image quality: (1) frequency filtering to remove object-like images, (2) patch variance filtering to eliminate homogeneous images, and (3) CLIP score filtering to ensure alignment with the prompts.

\subsubsection{Frequency Filtering for Object Removal}

The first step in our refinement process aims to remove images with object-like features by filtering based on frequency content. Object-based images typically consist of lower-frequency components, whereas textures generally exhibit higher-frequency patterns. To quantify this distinction, we compute the frequency profile of each image using Fourier transforms. Specifically, for each image, we calculate its Fourier transform and derive the power spectrum by examining the magnitude of each frequency component.

To characterize the frequency distribution, we define a \textit{frequency cutoff threshold}, \( f_c \), as the frequency where the cumulative energy from the power spectrum is split evenly between lower and higher frequencies. This threshold \( f_c \) is obtained by finding the frequency radius \( c \) at which:

\[
\sum_{k = 0}^{c} P(k) \approx \frac{1}{2} \sum_{k = 0}^{k_{\text{max}}} P(k)
\]

\noindent where \( P(k) \) represents the power at each frequency \( k \), and \( k_{\text{max}} \) is the maximum frequency band. A lower \( f_c \) indicates that more energy is concentrated in lower frequencies, which is characteristic of object-like images rather than textures.

From here, we retain only the top 80\% of images with the highest \( f_c \) values, ensuring that images with a higher concentration of high-frequency content are retained. This filtering is applied in a balanced manner across each texture class in the dataset. This frequency filtering step effectively removes images with prominent object-like characteristics while preserving the natural high-frequency patterns typical of textures.

\subsubsection{Patch Variance Filtering}
To eliminate homogeneous or near-uniform images, we apply a patch variance filtering step. We divide each image into non-overlapping patches of 50x50 pixels and calculate the mean intensity of each patch. We then compute the variance of these mean intensities across all patches in an image. Images with a low variance across patches are likely to lack textural detail, appearing overly smooth or homogeneous. We filter out the bottom 20\% of images with the lowest mean patch variance within each texture class, ensuring that PTD retains diverse textures with well-defined patterns and structures.

\subsubsection{CLIP Score Filtering for Prompt Alignment}
In the final refinement stage, we use CLIP-based filtering to ensure that the remaining images are well-aligned with their descriptive prompts. CLIP scores, which measure cosine similarity between text and image embeddings, provide a metric for representativeness. For each texture class, we filter out the bottom 20\% of images with the lowest CLIP scores, retaining those that most closely align with the intended descriptors. This step ensures that the final dataset reflects the prompt-based generation criteria for each texture class.

\subsubsection{Final Dataset}
After applying these three filtering stages, our refined Prompted Textures Dataset (PTD) contains \imgcountfinal{} images, balanced across all texture classes and meeting our quality standards for textural content and prompt alignment. This multi-stage refinement process enables PTD to serve as a high-quality, representative dataset for texture-based research and applications.

\section{Quality evaluation}\label{sec:quality_eval}

With our final Prompted Textures Dataset (PTD), totaling \imgcountfinal{} images, we evaluate the quality, diversity, and representativeness of the dataset in two parts: (1) through a comparison of Inception and FID scores with an existing texture dataset, the Describable Textures Dataset (DTD), and (2) with a human evaluation study.

Image generation and quality evaluation experiments were run on 12 A100 GPUs with 40GB of memory each using CUDA version 11.8. Images are generated using the Stable Diffusion model~\cite{rombach_high-resolution_2022} from HuggingFace~\cite{wolf_huggingfaces_2020}. Our dataset is available for download at \url{https://zenodo.org/records/15359142}.

\subsection{Standard Metrics}\label{sec:qual:standard}

Inception~\cite{salimans_improved_2016} and FID~\cite{bynagari_gans_2019} scores are standard metrics commonly used to assess the quality and diversity of generated image datasets. Inception scores reflect both the image quality and diversity within the dataset, with higher scores indicating better performance, while FID scores measure the similarity between the generated dataset and a real dataset, with lower scores indicating closer alignment. Although both metrics were originally designed with object-based datasets in mind, they can still provide insight into the representativeness and quality of a texture dataset when compared to a real-world baseline, such as DTD.

\begin{figure*}[t]
  \centering
  \includegraphics[width=\textwidth]{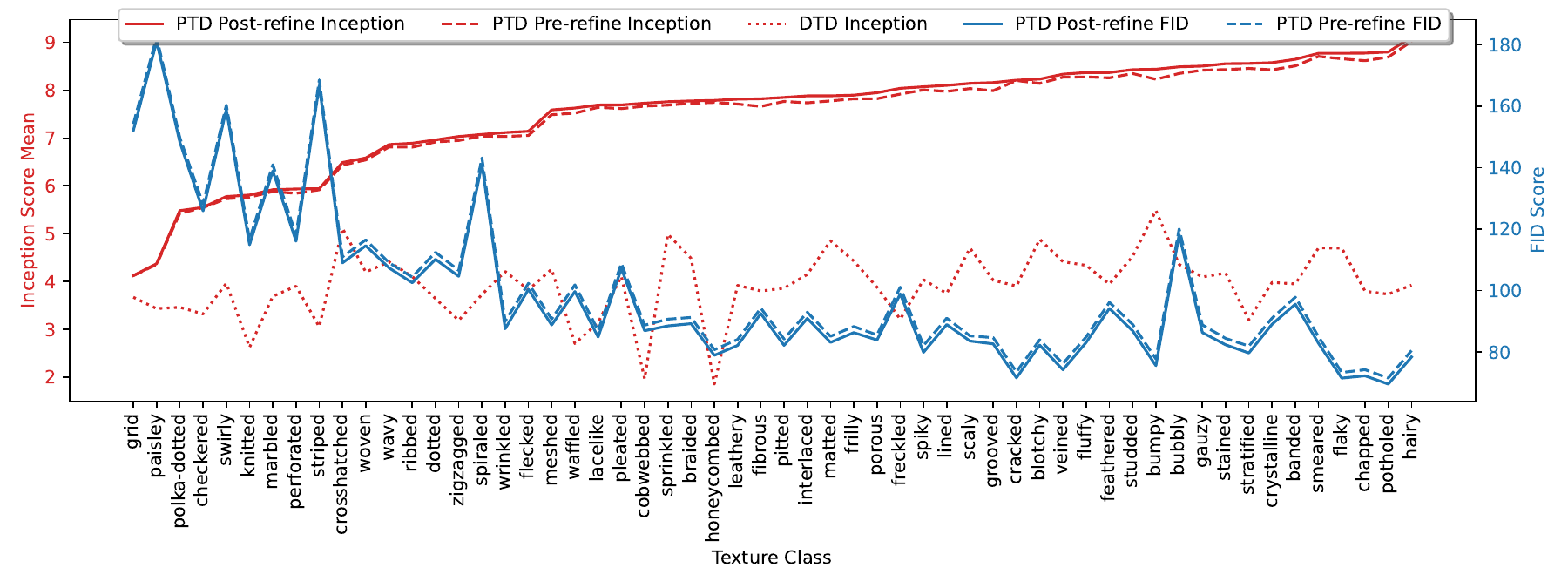}
  \caption{Inception and FID Scores of each texture class in PTD (ours) and DTD\cite{cimpoi_describing_2014}. Classes are sorted by mean Inception Score.}
  \label{fig:initial_quality_overlay}
\end{figure*}

We evaluate PTD, both pre- and post-refinement, using both metrics to determine overall comparison to DTD and also how well our refinement process has improved the dataset’s quality according to standard metrics. In \autoref{fig:initial_quality_overlay}, we report Inception and FID scores, separated by texture class, for PTD both before and after refinement, with DTD scores included for direct comparison. Note that, unlike the Inception scores, the FID scores incorporate both the PTD and DTD data directly, so the "PTD pre-refine FID" and "PTD post-refine FID" values measure the similarity between the DTD with the PTD pre-refinement and post-refinement, respectively. From these results, we find three interesting trends.



\noindent\textbf{Effective Refinement Process:} Our refinement process improves both the Inception and FID scores for PTD, indicating that the multi-step filtering effectively enhances the dataset’s quality. Post-refinement Inception scores are higher, and FID scores are lower across all texture classes, showing that the filtered dataset is not only more diverse and high quality but also more similar to the realistic textures found in DTD. This confirms that the combination of frequency-based, patch variance, and CLIP filtering steps successfully curates PTD to remove non-representative images while preserving textural complexity.

\noindent\textbf{PTD Outperforms DTD in Diversity:} Inception scores for PTD consistently surpass those of DTD across texture classes, suggesting that PTD is more diverse and thus offers broader potential for future research in texture-based tasks. This diversity likely arises from the varied and descriptive prompts used during generation, which cover a wider range of textural characteristics compared to DTD. 

\noindent\textbf{Variation Across Texture Classes:} While overall scores showcase the quality and diversity of PTD, we observe variations in both Inception and FID scores across texture classes, with certain textures performing better than others. Notably, textures that more often appear in common objects, such as \textit{hairy} and \textit{flaky}, despite being more complex textures, tend to have better scores. Meanwhile, more simple, but more structured textures, such as \textit{grid} and \textit{polka-dotted} textures tend to have worse scores. This indicates that while the refinement process enhances general quality, there is room for improvement in generating more structured textures.

Overall, these findings confirm that the PTD is a high-quality, diverse dataset with strong alignment to DTD, while also offering an expanded range of textures. To further validate these insights, we complement these quantitative metrics with a human evaluation study to verify perceptual quality and representativeness.

\subsection{Fourier Analysis}

As a final automated quality evaluation before our human evaluation, we also perform a Fourier analysis on our
dataset. This analysis is used to determine the frequency of the textures in our
dataset, and to ensure that the textures are not overly biased towards a certain
frequency. This analysis is important because it can help to ensure that the
textures in our dataset are diverse and not overly biased towards a certain type
of texture. 

\begin{figure}[t]
  \includegraphics[width=\linewidth]{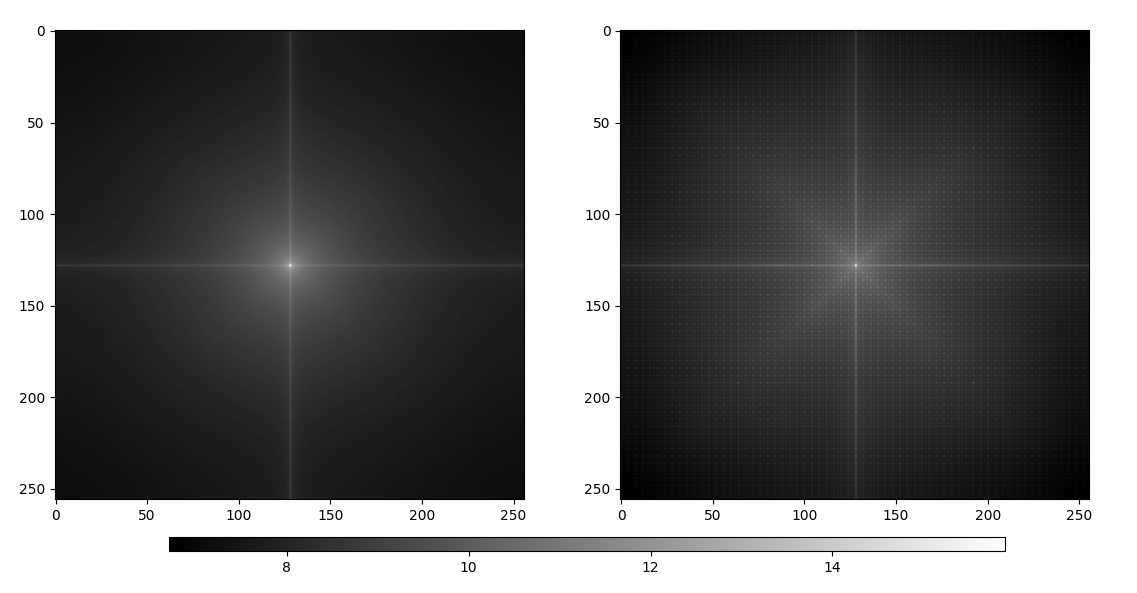}
  \caption{Mean power spectrum of DTD (left) and PTD (right).}
  \vspace{0.4cm}
  \label{fig:fourier}
\end{figure}

To perform this analysis, we first convert the images in our dataset to their
Fourier representations. We then calculate the power spectrum of the images,
which gives us the frequency of the textures in the images. We then calculate
the mean power spectrum of the images in our dataset, and compare this to the
mean power spectrum of the images from the Describable Textures
Dataset (DTD) \cite{cimpoi_describing_2014}. The results of this analysis are
shown in \autoref{fig:fourier}. From the similarity between these two images, we can see that the textures in
our dataset are not biased towards a certain frequency and well capture the
frequencies found in real data (e.g., the DTD). This is a good
indication that the textures in our dataset are diverse and representative of a
wide range of textures. 


\subsection{Human Evaluation}\label{sec:human_eval}

To further validate the quality of the Prompted Textures Dataset (PTD) and assess the effectiveness of our refinement steps, we conduct a human evaluation on the images in our dataset. For our human evaluation, we recruited 9 participants to evaluate the images. Each participant was shown 100 images in random order and asked two questions for each image: (1) How would you rate the overall \textbf{quality} of the image? and (2) How well does the image \textbf{represent} the provided descriptor? Participants were asked to supply a rating on a scale of 1 to 5, with 1 being the worst and 5 being the best, for each of these questions for every image. 

The images for the image sets provided to the participants were selected randomly from the dataset, but we ensured there were no duplicate images between or within the sets, meaning that we evaluated 900 unique images from our dataset. These images were selected before the refinement stage in our pipeline, such that some of the evaluated images were removed as part of our refinement process. This was done to compare the human evaluation scores before and after refinement to see if our refinement process does indeed help to improve the overall quality of our dataset. 

\subsubsection{Image Quality}

In \autoref{tab:human_eval} we show the results of the human evaluation. Here, we take the mean quality and representative scores, provided by our human evaluators, across the images at different stages of the refinement process. At each refinement stage, the current refinement process is applied in addition to all refinement steps that come before it (e.g., the \textit{Patch Var} refinement step also includes the \textit{Freq} refinement). From these results, we observe a few trends.

\begin{table}[t]
\centering
\begin{tabular}{l|cc}
\toprule
Refinement Step & Quality & Representative \\
\midrule
None & 3.87 & 3.56 \\
+Freq & 3.89 \textcolor{blue}{(+0.02)} & 3.58 \textcolor{blue}{(+0.02)} \\
+Patch Var & 3.95 \textcolor{blue}{(+0.06)} & 3.63 \textcolor{blue}{(+0.05)} \\
+CLIP & 4.00 \textcolor{blue}{(+0.05)} & 3.72 \textcolor{blue}{(+0.09)} \\
\bottomrule
\end{tabular}
\caption{Mean human quality and representativeness scores for each refinement step, with improvement shown in blue.}
\label{tab:human_eval}
\end{table}

\noindent\textbf{Gradual Improvement through Refinement:} At each step in the refinement process, we can see that both quality and representativeness of the data improves from the previous step, confirming that each part of our refinement process leads to improved dataset quality. Combining all refinement steps together provides us with an overall 3.4\% increase in quality and a 4.5\% increase in representativeness.

\noindent\textbf{Effectiveness of Patch Variance and CLIP:} Observing the increases at each step, we see that Patch Variance and CLIP refinement contribute the most to the quality and representative score increases, respectively. For CLIP scores, this result orients well with the goal of the metric, which measures the alignment between prompt and images, and serves as confirmation of prior results that CLIP scores can well represent human evaluators. For Patch Variance refinement, the increase suggests that human evaluators tend to favor textures that are less homogeneous and structured, providing new insights into properties of high-quality textures.

\subsubsection{Quality trends across prompts}\label{sec:human_eval:prompt}

\begin{figure}[t]
  \centering
  \includegraphics[width=0.8\linewidth]{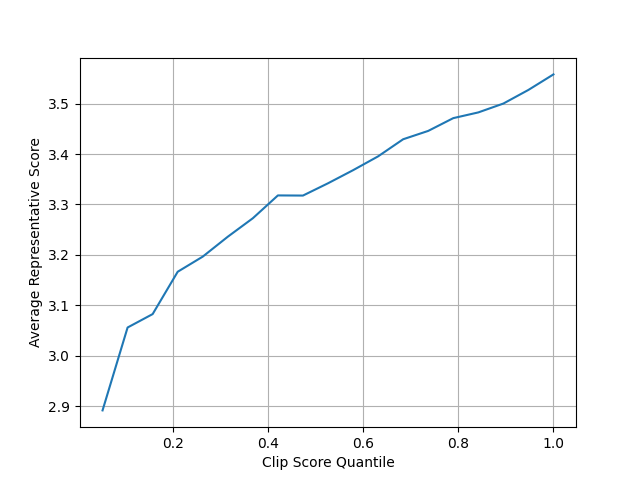}
  \caption{Average human representative score for all images at or below a given CLIP score quantile cutoff.}
  \vspace{0.3cm}
  \label{fig:clip_vs_human}
\end{figure}

In addition to the overall quality of the images in PTD, we additionally aim to understand what kinds of prompts result in the generation of high-quality images, which can help inform future works on prompt generation for textural data or extensions to the PTD. In other words, here we explore the question \textit{how do different descriptors affect the quality of the images produced?} We analyze these prompt trends using the representative scores assigned by human evaluators, prompt-specific commentary provided by human evaluators, and CLIP scores. Given the alignment we observed in the previous results between human assigned representative scores and CLIP scores, which we further validate in \autoref{fig:clip_vs_human}.

Table \ref{tab:clip_table} shows the top and bottom 5 mean CLIP scores across all descriptor word pairs. 
Among the prompt pairs that do tend toward the top or bottom, we see some word pairing clusters. The texture ``woven'' appears to lead to higher quality images when paired with basic colors such as red, green, and blue. In contrast, more subtle textures such as ``gauzy'' and ``veined'' seem to result in lower CLIP scores when paired with descriptors that are designed to make the image more subtle, such as ``muted'' and ``earthy''. From this, we find that some word pairs are more compatible than others and that this can influence resulting image quality.

\begin{table}[t]
\centering
\begin{tabular}{llll}
\toprule
Word pair & Mean & Median & \# Samples \\
\midrule
woven blue & 29.52 & 29.84 & 1080 \\
woven red & 29.49 & 29.86 & 1080 \\
marbled photo-realistic & 29.48 & 29.71 & 2160 \\
woven green & 29.47 & 29.91 & 1080 \\
woven $\emptyset$(\textit{color enhancer}) & 29.37 & 29.5 & 960 \\
... & ... & ... & ... \\
frilly neutral-colored & 23.74 & 23.88 & 1080 \\
veined earthy & 23.68 & 24.05 & 960 \\
gauzy earthy & 23.51 & 23.86 & 960 \\
gauzy muted & 23.28 & 23.62 & 960 \\
veined muted & 23.2 & 23.35 & 960 \\
\bottomrule
\end{tabular}

\caption{Top and bottom 5 CLIP scores across descriptor word pairs.}
\label{tab:clip_table}
\end{table}

From the human evaluation, in addition to the raw scores provided on the images, the participants also had the option of commenting on any trends they may have observed when evaluating the images. The most common comments we observed were: the descriptors containing the word ``muted'' often had ``destroyed'' images in terms of quality (P1), describing colors often alters the background (P2), symmetric images are sometimes not symmetric (P2), colors are very well represented in the images (P3), and descriptions with fewer words looked more realistic (P4).

These comments also agree with our results on assessing prompt quality. In particular, when looking at prompt pairs that generated images with higher or lower mean CLIP scores, we find that ``muted'' is one example of a descriptor whose images tend to be toward the bottom of the mean CLIP scores. For example, both \textit{veined muted} and \textit{gauzy muted} had mean CLIP scores of 23.20 and 23.28, respectively. 
These results demonstrate that CLIP can effectively represent human scores for alignment between prompts and images, even on texture data. Analyzing both human scores and comments, and CLIP scores, we find combinations of descriptors that yield higher and lower quality images, providing insight to future work on texture generation.




\section{Measuring Texture Bias}

In this section, we highlight the usefulness of the Prompted Textures Dataset as a way to measure texture bias. \textit{Texture bias} refers to a model's affinity to learn, and be biased towards, texture information rather than shape information~\cite{geirhos_imagenet-trained_2019}. This phenomenon has served as a puzzling and highly interesting discovery primarily due to the fact that human vision is more biased towards shape information rather than texture information, and the impressive performance of computer vision models would suggest that they would learn similar information. The initial discovery of texture bias has spurred numerous subsequent works that work toward understanding why models are bias towards texture~\cite{hermann_origins_2020}, but until recently there has only been one standard method for measuring a model's bias towards texture information~\cite{geirhos_imagenet-trained_2019}, and this approach has been limited by size and scope of the texture data (due to the burden of manual collection) and assumptions on textures learned by models.

Since the initial release of the Prompted Textures Dataset, it has reached 23 downloads on Zenodo (link excluded for anonymization) at the time of submitting this paper and has been used as the forefront for developing new texture bias measurement methodologies. Most recently, the Prompted Textures Dataset served as the basis for \textit{Texture Object Association Values (TAV)}~\cite{hoak_err_2024}, a data-driven metric that computes the level of associativity or relatedness of textures with objects by analyzing model responses to the Prompted Textures Dataset. By leveraging the Prompted Textures Dataset, the new TAV metric was able to identify real textures present in images and subsequently measure texture bias by quantifying how predictions changed in the presence of different textures. From this, it was found that the texture type can affect the accuracy (and confidence) of the model by up to 66\% (and 40\%) on clean validation data, demonstrating that accurate and confident predictions rely on the presence of certain textures. Furthermore, it was found that 90\% of the natural adversarial examples contained misaligned texture information with their true class, which explained their confidently incorrect classifications~\cite{hoak_err_2024}. \textbf{These new metrics and findings open up new possibilities towards studying texture bias and its impact on model trustworthiness, and would not have been possible without the Prompted Texture Dataset that we introduce in this work.}  

\begin{figure}
    \centering
    \includegraphics[width=\linewidth]{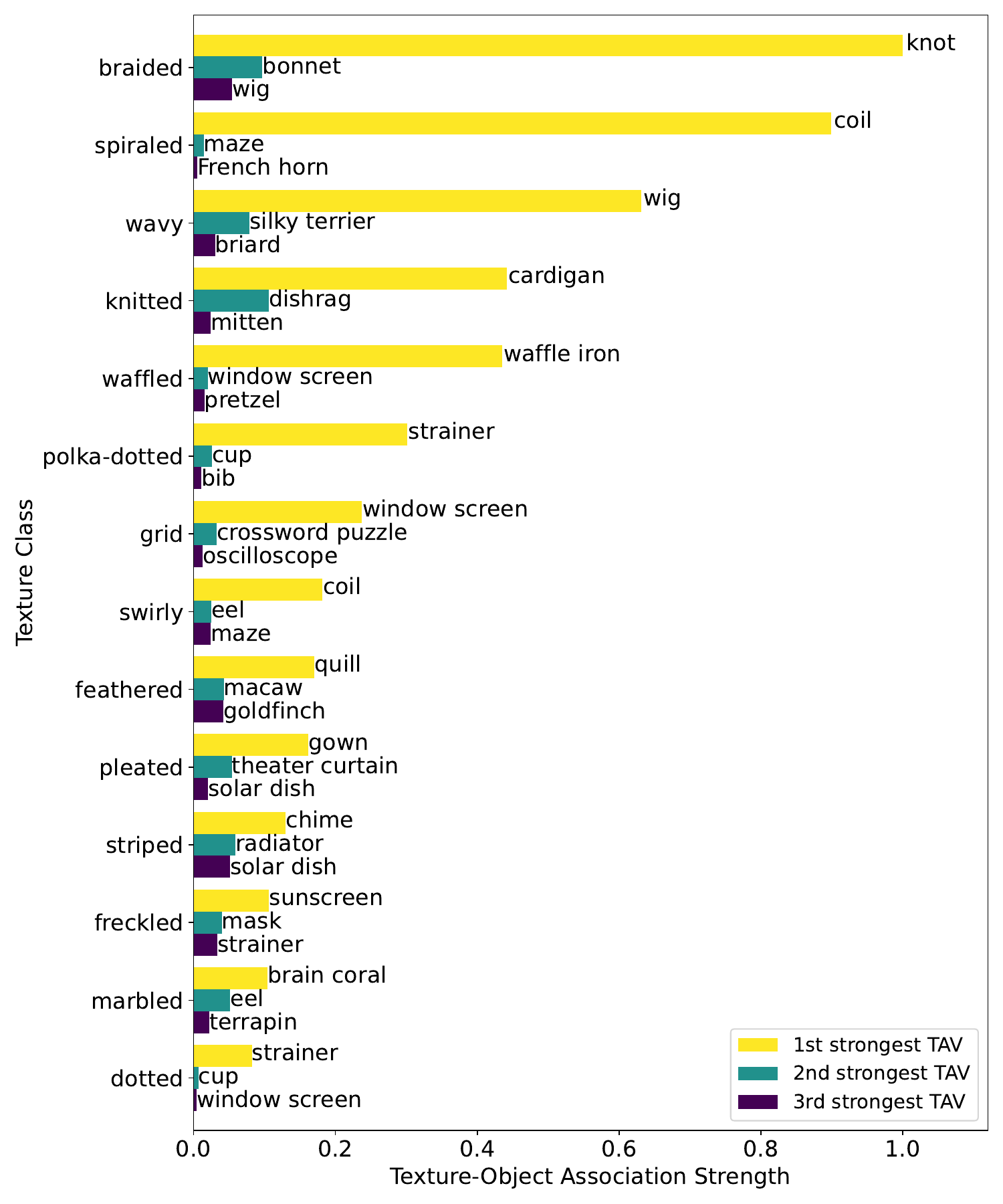}
    \caption{Top 3 highest Texture Object Associations~\cite{hoak_err_2024} on ResNet 50 using the Prompted Textures Dataset.}
    \label{fig:text_obj}
\end{figure}

To showcase the utility of the Prompted Textures Dataset, we implement the Texture Object Association Value (TAV) metric as introduced in Hoak et al.~\cite{hoak_err_2024}, which uses the Prompted Textures Dataset to uncover learned associations between objects and textures. In \autoref{fig:text_obj}, we show the resulting association values for 14 of our texture classes on ResNet50. Each set of 3 bars shows the 3 highest association values for the given texture class, annotated with the associated object class. From these association results, we see that the Prompted Textures Dataset can uncover meaningful associations that identify the textures the model relies on when classifying various objects. Here we see that some of the strongest associations are braided textures with knot objects, spiraled textures with coil objects, and wavy textures with wig objects. The fact that these associations are ones that make logical sense to humans as well demonstrates that the Prompted Textures Dataset is effective in uncovering texture-object associations and thus measuring texture bias. 

\section{Conclusions}
In this work, we presented a novel methodology for generating high-quality, diverse texture images, addressing key limitations in the diversity and scalability of other texture datasets and synthesis methods. By leveraging text-to-image models and incorporating tailored prompt design along with a three stage refinement process, we created the Prompted Textures Dataset (PTD) to enable new exploration in texture-based research. Our evaluations reveal that PTD not only meets high standards of quality and diversity but also surpasses the representational capabilities of traditional datasets like DTD. Through human evaluations, we confirmed the effectiveness of each refinement step, enhancing both the quality and representativeness of the images. Finally, we analyzed the use of the Prompted Texture Dataset in newly developed methods for measuring texture bias and found that the PTD uncovers useful and sensible texture-object associations, and thus is an effective dataset for measuring texture bias. This work demonstrates that generative models, when carefully adapted, can yield extensive, versatile datasets for texture analysis, contributing valuable resources for advancing interpretability, bias analysis, and robustness in machine learning. 



\begin{ack}
This material is based upon work supported in part by the
National Science Foundation under Grant No. CNS-2343611
and in part by the U.S. Army Research Office under Grant
W911NF2110317. Any opinions, findings, and conclusions
or recommendations expressed in this material are those of
the author(s) and do not necessarily reflect the views of the
National Science Foundation.
The authors would also like to thank all the participants
that made our human evaluation possible, as well as Eric
Pauley for his feedback on early drafts of the paper and
experiment design.
\end{ack}



\bibliography{references}

\end{document}